\documentclass[article, 10 pt, conference]{IEEEtran}

\usepackage{times}
\usepackage{soul}
\usepackage{url}
\usepackage[hidelinks]{hyperref}
\usepackage[utf8]{inputenc}
\usepackage[small]{caption}
\usepackage{graphicx}
\usepackage{amsmath}
\usepackage{booktabs}
\usepackage[font=footnotesize]{caption}
\usepackage{cite}
\usepackage{enumitem}

\setlist[itemize]{noitemsep, topsep=0pt}

\title{\LARGE \bf Mixing Human Demonstrations with Self-Exploration in Experience Replay for Deep Reinforcement Learning}

\author{
Dylan Klein and Akansel Cosgun\\
Monash University, Australia\\
}

\begin{document}

\maketitle

\begin{abstract}
We investigate the effect of using human demonstration data in the replay buffer for Deep Reinforcement Learning. We use a policy gradient method with a modified experience replay buffer where a human demonstration experience is sampled with a given probability. We analyze different ratios of using demonstration data in a task where an agent attempts to reach a goal while avoiding obstacles. Our results suggest that while the agents trained by pure self-exploration and pure demonstration had similar success rates, the pure demonstration model converged faster to solutions with less number of steps.
\end{abstract}

\section{Introduction}

Deep Reinforcement Learning (DRL) is a promising method for sequential decision making and is used in different platforms including robotic manipulators \cite{lee2020learning}, drones \cite{de2021decentralized} and mobile robots \cite{tidd2021passing}. Despite its promise, DRL struggles for large state spaces and sparse rewards. 

Exploration for reinforcement learning is often driven by stochastic processes, however there has been recent efforts to use human demonstrations for overcoming exploration problems~\cite{subramanian2016exploration,nair2018overcoming}.~\cite{gao2018reinforcement} proposes an actor critic method that initializes the policy network from demonstrations and refines it with self-exploration.~\cite{hester2018deep} showed that a DQN trained on a combination of demonstration  and self-exploration performed better than pure self-exploration in Atari games.

Replaying past experiences in machine learning was first explored by~\cite{lin1992self}. In deep reinforcement learning, an experience replay buffers is commonly used to store past experiences~\cite{isele2018selective,mnih2015human}, and often implemented as First-in-First-Out (FIFO) buffers where experiences are sampled randomly.


In this work we compare the performance achieved by selecting experiences from human demonstration data and self-exploration. We base our work on modifying the experience replay buffer of a variant of the Deep Deterministic Policy Gradient (DDPG) algorithm~\cite{lowe2017multi}. We store the data from self-exploration and demonstrations separately in the experience replay buffer, which allows us to adjust the proportion of the sampled experiences. We compare the performances of using pure exploration, pure demonstration and a mixture of the two. A simulated navigation problem is used for evaluation where the agent's goal is to reach a goal while avoiding obstacles.




\section{Mixed Experience Replay}
\label{sec:mixed_experience_replay}

We first train a pure self-exploration policy and capture 100,000 experiences. Separately, human demonstrations were performed in order to capture an additional 100,000 experiences. Each set of 100,000 experiences were saved in separate buffers. An additional experience replay buffer was then created by combining the experiences that were captured using pure self-exploration with those of the demonstrations. The additional experience replay buffer was populated by use of a sampling probability $p$ which dictated the likelihood of the next experience being sampled from the demonstrations buffer. Thus this experience replay buffer held experiences which were derived from mixed types of training, both pure self-exploration and demonstrations. Increasing or decreasing the value of $p$ therefore controlled the relative proportion of pure self-exploration experiences to demonstrations in the mixed experience replay buffer.

\vspace{-0.2cm}
\section{Experiments}
\label{sec:experiments}
\begin{figure}[ht!]
    \centering
    \includegraphics[width=0.5\linewidth]{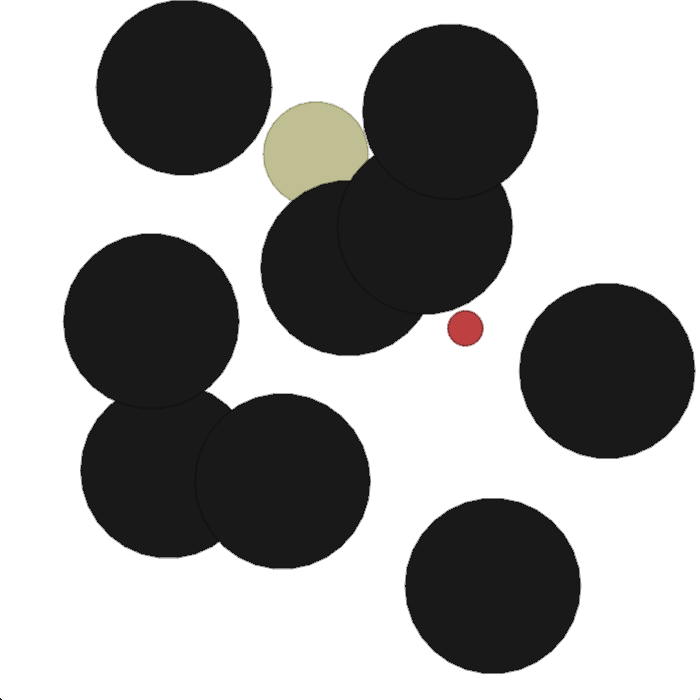}
    \caption{A snapshot of the 2-dimensional environment including the actor (in gold), obstacles (in black) and goal (in red).}
    \label{fig:snap}
\end{figure}

An actor was placed in a 2-dimensional environment which included 9 path-blocking obstacles, as seen in Figure~\ref{fig:snap}. The agent was rewarded by minimizing its Euclidean distance to the goal. State information was encoded as a 22-dimensional vector which included the $x$ and $y$ components of the actor's velocity, the actor's relative position to its goal and the actor's relative position to each of the 9 obstacles. The action space was encoded as a 5-dimensional vector containing quantities between 0 and 1, which indicated the actor's incremental positional change. For pure self-exploration, the quantities were continuous values. Whereas for demonstrations, the quantities were discrete binary and inputted via the keyboard. At each time step the agent receives a reward equal to minus distance to the goal.

When pure self-exploration experiences were captured, the actor was a RL agent with the following function:

\begin{itemize}
\setlength\itemsep{0pt}
    \item Input: A pure self-exploration experience replay buffer, containing state, action, reward, next step's state and episode termination information.
    \item Output: An updated pure self-exploration experience replay buffer, containing state, action, reward, next step's state and episode termination information.
\end{itemize}

When demonstrations were performed, the human was the actor with the following function:

\begin{itemize}
\setlength\itemsep{0pt}
    \item Input: A visual display of the game screen.
    \item Output:  A pure human-demonstration experience replay buffer, containing state, action, reward, next step's state and episode termination information.
\end{itemize}

\section{Results}
\label{sec:results}
A neural network was trained with the mixed experience replay buffer as its input. 10 models of the neural network were saved every 100 episodes (1000 episodes in total) as outputs. Each model was independently evaluated against 2 task performance metrics:

\begin{itemize}
\setlength\itemsep{0pt}
    \item Success rate: how often the agent reached its goal, as a percentage.
    \item Solution efficiency: the mean number of steps the agent took to reach its goal (for successful runs).
\end{itemize}

\begin{figure}[t!]
    \centering
    \includegraphics[clip, trim=0.2cm 0.2cm 0.2cm 0.2cm, width=1\linewidth]{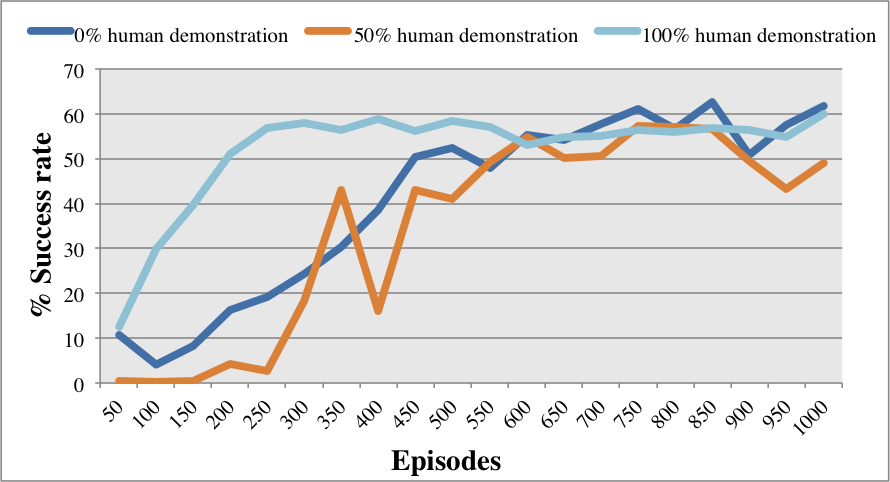}
    \caption{Success rate - percentage success rate for the agent to reach its goal during evaluation.}
    \label{fig:sr}
\end{figure}

Figure~\ref{fig:sr} indicates that a model which was trained from experiences that were controlled 100\% by demonstrations converges faster than pure self-exploration and mixed training models. This suggests that demonstration-based training is effective in narrowing down the space of possible solutions. It is worth noting that 0\% demonstration models and 100\% demonstration models reached to similar steady state values. The success rates were slightly above 60\% for both methods. A low success rate was achieved because sometimes there was no path to the goal due to random placement of obstacles.

The model trained by 50\% by demonstrations and 50\% by self-exploration has a far more volatile learning curve than that of pure self-exploration and pure demonstrations. This may suggest that mixed training tends to confuse an agent due to conflicting prescribed actions given a particular observed state.

\begin{figure}[ht!]
    \centering
    \includegraphics[clip, trim=0.2cm 0.2cm 0.2cm 0.2cm, width=1\linewidth]{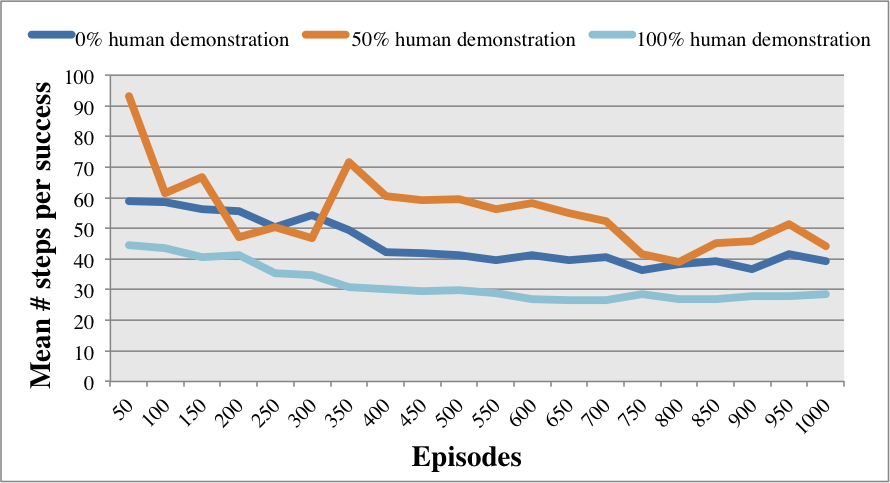}
    \caption{Solution efficiency - mean number of steps for the agent to reach its goal during evaluation.}
    \label{fig:se}
\end{figure}

Figure~\ref{fig:se} shows that 100\% demonstrations lend itself to finding more efficient solutions. In other words, less steps were required for an agent to reach its goal when a human previously provided examples of how to navigate efficiently. This aligns with the fast convergence of Figure~\ref{fig:sr} where demonstrations help in narrowing down the space of efficient solutions. Despite showing improvement over 1000 episodes, models trained from 0\% demonstrations did not appear to find solutions that were as efficient as those trained from 100\% demonstrations. Again, models trained from 50\% demonstrations under-performed all other models from a solution efficiency perspective.

\vspace{-0.2cm}
\section{Conclusion}
\label{sec:conclusion}

In this work we study the effect of using human demonstration data in the experience replay buffer. We compare three experience sampling strategies: pure self-exploration, pure demonstration and 50\% demonstration. In a simulated path finding scenario, we compared the approaches by according to two task metrics: the rate which the agent reaches the goal, and the number of steps taken when it does. The agents trained by pure self-exploration and pure demonstration had similar success rates at steady state. Pure demonstration strategy converged faster than the other strategies and it resulted in shorter paths to the goal. The 50\% demonstration strategy performed the worst of the three in both metrics.

A limitation of this study was that pure self-exploration implemented a continuous action space whereas human demonstrations implemented a discrete action space. This may explain poor performance of the policy at heavily mixed levels of training types, for example at 50\% demonstration.

\bibliographystyle{IEEEtran}
\bibliography{ijcai19}

\end{document}